\documentclass[11pt]{article}

\usepackage{dialogue}

\usepackage{ifxetex}
\ifxetex
    \usepackage{fontspec}
    \setromanfont{Times New Roman}
\else
  \usepackage{cmap}
  \usepackage[T2A,T1]{fontenc}
  \usepackage[utf8]{inputenc}
  \usepackage{times}
  \usepackage{latexsym}
  \DeclareFontFamilySubstitution{T2A}{\familydefault}{cmr}
\fi

\usepackage[russian,british]{babel}
\usepackage{url}
\usepackage{pgf}

\usepackage{longtable}
\usepackage{float}
\usepackage{covington}
\usepackage{hyperref}

\dialogfinalcopy

\title{\textbf{Ossetic-COT: Designing a morphologically annotated corpus and morphological analyzer for Ossetic}}

\author{Anna Shatskikh \\
  Lomonosov Moscow State University \\
  {\tt anapriselec@gmail.com} 
   \\\And
  Alexey Sorokin \\
  Lomonosov Moscow State University \\
   Yandex \\
  {\tt alexey.sorokin@list.ru} \\}

\date{}

\begin{document}
\maketitle
\begin{abstract}
  In this work we present the first morphologically annotated corpus for Iron Ossetic that conforms to the Universal Dependencies schema. The corpus includes 5454 manually annotated sentences from the Iron Ossetic Corpus of Oral Texts, containing 74032 tokens. We use this corpus to train a BERT-based morphological analyzer. The analyzer achieves tag accuracy of 95.60\%.
  
  
  \textbf{DOI:} 10.29003/2075-7182-2026-24-544-555
\end{abstract}

\selectlanguage{russian}
\begin{center}
  \russiantitle{Ossetic-COT: Создание корпуса с морфологической разметкой и морфологического анализатора для осетинского языка}

  \medskip \setlength\tabcolsep{2cm}
  \begin{tabular}{cc}
    \textbf{Анна Шатских} & \textbf{Алексей Сорокин}\\
      МГУ им. Ломоносова & МГУ им. Ломоносова, Яндекс \\
      {\tt anapriselec@gmail.com} & {\tt alexey.sorokin@list.ru} 
  \end{tabular}
  \medskip
\end{center}

\maketitle

\begin{abstract}
  Работа посвящена созданию первого корпуса для осетинского языка с морфологической разметкой в формате Universal Dependencies. Корпус состоит из размеченных вручную текстов Устного корпуса осетинского языка и содержит 5454 предложения суммарной длиной 73 042 токена. Также в работе представлен морфологический анализатор, основанный на модели BERT и достигающий пословной точности 95,60\%.
  
  \textbf{Ключевые слова:} осетинский язык, иронский диалект, Universal Dependencies, морфологический анализ
\end{abstract}

\selectlanguage{british}

\section{Introduction}
\label{sec-intro}
Ossetic is an Iranic language from the Indo-European family spoken by up to 641,450 speakers mostly in North Causasus region \footnote{http://www.ethnologue.com/show\_language.asp?code=oss}. There exist two main corpora of Ossetic: Ossetic National Corpus (ONC) \footnote{http://corpus.ossetic-studies.org} and the Corpus of Oral Texts (COT) \footnote{https://www.ossetic-studies.org/ru/texts/iron}.

ONC contains Ossetic literary texts, mostly written by contemporary authors (1999-2014), as well as the works by the most celebrated writers of the 20th century. It also includes texts from \foreignlanguage{russian}{<<Мах дуг>>} (<<Our epoch>>) and \foreignlanguage{russian}{<<Ногдзау>>} (<<Pioneer>>) literary magazines and \foreignlanguage{russian}{<<Спутник>>} (<<Satellite>>) online newspaper, as well as the Nart Epic collection. The size of ONC is approximately 12 millions of tokens. The corpus is automatically annotated for parts of speech, morphological features and lemmas via Uniparser-Morph \cite{archangelsky:12}. Because the parser is context-free, the corpus is not disambiguated, providing multiple possible tags for a given word.

COT includes Ossetic oral texts collected during linguistic expeditions in 2007-2013 in various regions of North Ossetia. It also contains a number of issues of \foreignlanguage{russian}{<<Одноклассники>>} (<<Classmates>>) radio program. The size of COT is 60,000 tokens. The corpus is provided with manual interlinear glossing done using SIL FieldWorks Language Explorer \footnote{https://software.sil.org/fieldworks}. 

Although it is possible to extract part-of-speech tags and morphological features from SIL format, the corpus has only incomplete coarse-grained annotation (see Section \ref{sec-guidelines}). This not only complicates automated conversion to the generally accepted Universal Dependencies annotation schema \cite{nivre:2020} but also limits the applicability of the corpus for educational or research purposes. These factors necessitate the development of a morphologically disambiguated Ossetic corpus that adheres to Universal Dependencies (UD) standards. The present work fulfills this objective.

The remainder of the paper is organized as follows. Section \ref{sec-background} provides necessary background linguistic information about Ossetic. In Section \ref{sec-method} we describe the corpus and annotation methodology. In Section \ref{sec-guidelines}, we examine several distinct morphological properties of Ossetic and describe the approach taken for their annotation. In Section \ref{sec-model} we describe how the corpus is used to train and evaluate a BERT-based morphological analyzer. Section \ref{sec-lim} describes the limitations of the corpus and the Section \ref{sec-concl} concludes.

\section{Background}
\label{sec-background}
Ossetic has two main dialects, Iron and Digor. In this paper, we use Ossetic and Iron Ossetic interchangeably.

Ossetic has a rich morphological system. Its nominal morphology is agglutinative with regular vowel and consonant alternations in plural formation \cite{abaev:64}. The Ossetic verbal system is highly synthetic, characterized by distinct inflectional classes, suppletive or irregular past stem formation, and the fused expression of person and number. The formation of complex predicates is highly productive. They are formed by combining a nominal element with a light verb selected from a restricted set.

The clausal word order of Ossetic is flexible, with SOV (subject-object-verb) serving as the unmarked baseline. Ossetic is consistently head final. Concerning NP-internal structure, it strictly follows the hierarchy Poss > Det > Num > Adj, Gen > N \cite{belyaev:14}.

\section{Data sources}

\label{sec-method}

The goal of the present study is to create a morphologically annotated corpus for Ossetic. We choose the Ossetic Corpus of Oral Texts (COT) as the source corpus because it already possesses morphological annotation in a custom format. We convert its morphological tags to CoNLL-U format, utilizing Universal Dependencies version 2 (UD v2) Annotation Guidelines \cite{nivre:2020} with language-specific extensions discussed in Section \ref{sec-guidelines}. 

The conversion workflow comprises the following stages:

\begin{enumerate}
    \item Filtering of non-target language content, specifically the exclusion of all Russian sentences from the corpus. 
    \item Manual sentence segmentation, including bisection of sentences containing ellipsis followed by an uppercase letter and manual segmentation of multi-word tokens. This stage includes case normalization as well.
    \item Manual mapping of part-of-speech tags and features to the Universal Dependencies inventory.
    In the original annotation, a subset of PronType values is represented as lexical categories for pronouns. We supplement this by enriching the pronominal entries from the previously underspecified categories with missing feature values. Furthermore, we incorporate the Num value into the Case feature and extend the inventory of non-finite forms by adding the Red (Reduplicated) category (see Section \ref{sec-guidelines}).
    \item Manual refinement of tags and features according both to universal and language-specific annotation guidelines. This stage, which requires specialized expertise in Ossetic morphology, proved to be the most labor-intensive part of the conversion process.
    \item Automated conversion to CoNLL-U format, including the addition of default feature values not present explicitly in the original interlinear glosses, e.g., Nominative for the case feature.
\end{enumerate}

The resulting dataset contains 5454 sentences with 73042 tokens. It is divided into train, validation and test sets with a ratio of 8:1:1. Frequency distribution of sentences belonging to each subgenre is reported in Appendix \ref{app-gen}.

\section{Annotation Guidelines}
\label{sec-guidelines}

Due to the specific linguistic idiosyncrasies of Ossetic, the annotation process necessitates exceeding the standard set of features and values defined by the UD framework. While the original corpus annotation already incorporates several language-specific extensions, we further augment it by introducing additional modifications. This section details the extensions adopted from the original scheme, as well as those implemented specifically for the current study.

\subsection{Substantives}
It has frequently been observed that Ossetic exhibits a lack of clear distinction between nouns, adjectives, adverbs, and postpositions \cite{abaev:64}. The same sequence of characters may function as multiple parts of speech; for instance, the word \foreignlanguage{russian}{\textit{хорз}} translates as both `good' and `well'. Due to this functional overlap, COT assigns a common part-of-speech tag to nouns, adjectives, postpositions, and, in certain instances, adverbs.

In contrast, this study aims to achieve a more fine-grained, context-dependent disambiguation by implementing the Universal POS inventory and adhering to UD v2 guidelines \cite{nivre:2020}. The disambiguation is based on the following morphosyntactic criteria:
\begin{enumerate}
    \item Substantives without case morphology functioning as verbal modifiers and governing a dependent in genitive case are categorized as ADP (e.g., \foreignlanguage{russian}{\textit{бын}} in example \ref{ex-adp}).
    \item Substantives without case morphology functioning as verbal modifiers without dependents are categorized as ADV (e.g., \foreignlanguage{russian}{\textit{рӕсугъд}} in example \ref{ex-adv}).
    \item Substantives without case morphology functioning as modifiers of other substantives are categorized as ADJ (e.g., \foreignlanguage{russian}{\textit{ирон}} in example \ref{ex-adj}).
    \item All the other substantives are categorized as NOUN.
\end{enumerate}
\selectlanguage{russian}
\begin{examples}
    \item \label{ex-adp}
            \selectlanguage{russian}
            \gll Ӕмӕ зӕй-ы бын фе-сты адӕм...
            \foreignlanguage{british}{and} \foreignlanguage{british}{avalanche-\textsc{gen}} \foreignlanguage{british}{under} \foreignlanguage{british}{\textsc{pv}-be.\textsc{prs.3pl}} \foreignlanguage{british}{person.\textsc{pl}}
            \glt \foreignlanguage{british}{`And under the avalanche there were people...'}
            \glend
    \item \label{ex-adv}
            \gll Ды тынг рӕсугъд дзур-ыс.
            \foreignlanguage{british}{\textsc{2sg}} \foreignlanguage{british}{very} \foreignlanguage{british}{nicely} \foreignlanguage{british}{speak-\textsc{prs.2sg}}
            \glt \foreignlanguage{british}{`You speak very nicely.'}
            \glend
    \item \label{ex-adj}
            \gll Фӕлӕ ирон ӕгъдӕу-тт-ӕ…
            \foreignlanguage{british}{but} \foreignlanguage{british}{ossetic} \foreignlanguage{british}{tradition-\textsc{pl}-\textsc{nom}}
            \glt \foreignlanguage{british}{`But, Ossetic traditions…'}
            \glend
\end{examples}

\selectlanguage{british}
In the case of complex predicates or nominal predication, where the aforementioned criteria are inapplicable, the \textsc{NOUN} tag is assigned by default. However, we interpret the presence of comparative morphology or intensifying modifiers as diagnostic of adjectival behavior. Consequently, such lexemes are categorized as \textsc{ADJ}, as illustrated in example \ref{ex-intens}, where the intensifier \foreignlanguage{russian}{\textit{тынг}} functions as a dependent of \foreignlanguage{russian}{\textit{кадджын}}.
\selectlanguage{russian}
\begin{examples}
    \item \label{ex-intens}
            \gll Ӕмӕ гъе уыдон ны-ххас-т-а, ӕмӕ уый тынг кадджын уыд-и.
            \foreignlanguage{british}{and} \foreignlanguage{british}{well} \foreignlanguage{british}{DemDist.\textsc{pl.gen}} \foreignlanguage{british}{\textsc{pv}-take-\textsc{tr-pst.3sg}} \foreignlanguage{british}{and} \foreignlanguage{british}{DemDist.\textsc{sg.nom}} \foreignlanguage{british}{very} \foreignlanguage{british}{honourable} \foreignlanguage{british}{be-\textsc{pst.3sg}}
            \glt \foreignlanguage{british}{`And he took them, and it was very honourable.'}
            \glend
\end{examples}

\selectlanguage{british}
\subsection{Auxilaries}
Ossetic possesses a closed class of verbs capable of functioning as auxiliaries; however, all members of this class may also serve as lexical verbs. Although the original COT annotation does not employ a distinct \textsc{AUX} tag, we claim that distinction between auxiliary and lexical uses is both methodologically necessary and feasible based on syntactic criteria. Consequently, we have restricted the AUX designation to a specific range of contexts, defined as follows:
\begin{enumerate}
    \item Light verbs within complex predicates (e.g., \foreignlanguage{russian}{\textit{кодтат}} in example \ref{ex-cp}).
    \item Copulas in nominal predications (e.g., \foreignlanguage{russian}{\textit{уыди}} in example \ref{ex-np}).
    \item Auxilary verbs (proper) in analytic constructions (e.g., \foreignlanguage{russian}{\textit{нӕй}} in example \ref{ex-ac}), including causative verb in causative constructions.
\end{enumerate}
\selectlanguage{russian}
\begin{examples}
    \item \label{ex-cp}
            \gll Куыд ахуыр код-т-ат скъола-йы?
            \foreignlanguage{british}{how} \foreignlanguage{british}{study} \foreignlanguage{british}{do-\textsc{tr-pst.2pl}} \foreignlanguage{british}{school-\textsc{ine}}
            \glt \foreignlanguage{british}{`How did you learn at school?'}
            \glend
    \item \label{ex-np}
            \gll Алагирский район-ы хицау уыд-и.
            \foreignlanguage{british}{Alagir} \foreignlanguage{british}{region-\textsc{gen}} \foreignlanguage{british}{head} \foreignlanguage{british}{be-\textsc{pst.3sg}}
            \glt \foreignlanguage{british}{`He was the head of Alagir region.'}
            \glend
    \item \label{ex-ac}
            \gll Уым дӕр нӕй фыст.
            \foreignlanguage{british}{DemDist.\textsc{ine}} \foreignlanguage{british}{\textsc{add}} \foreignlanguage{british}{be.\textsc{neg.prs.3sg}} \foreignlanguage{british}{write.\textsc{ptcp.pst}}
            \glt \foreignlanguage{british}{`It isn’t written there too.'}
            \glend
\end{examples}

\selectlanguage{british}
\subsection{Pronouns}
While UD v2 permits the lexical pre-classification of pronominal items as either PRON or DET based on their prototypical distribution and morphology, for Ossetic we adopt a context-based approach consistent with our treatment of NOUN and ADJ. This decision is motivated by the fact that lexemes typically functioning as determiners frequently undergo substantivation, appearing as verbal arguments or bearing nominal morphology (e.g., \foreignlanguage{russian}{\textit{ахӕмтӕ}} in example \ref{ex-det}). In such instances, the PRON tag is assigned, accompanied by Case and Number features. The same principles apply to the placeholder lexeme \foreignlanguage{russian}{\textit{йед}} (e.g. example \ref{ex-jed}). It should be noted, however, that this particular lexeme lacks the PronType feature, as it does not align with any of the categories established in the UD v2 framework.
\selectlanguage{russian}
\begin{examples}
    \item \label{ex-det}
            \gll Уыд-и дзы ахӕм-т-ӕ дӕр, ӕртын фондз аз-ы кӕ-уыл цыд-и.
            \foreignlanguage{british}{be-\textsc{pst.3sg}} \foreignlanguage{british}{\textsc{3sg.ine}} \foreignlanguage{british}{such-\textsc{pl-nom}} \foreignlanguage{british}{\textsc{add}} \foreignlanguage{british}{thirty} \foreignlanguage{british}{five} \foreignlanguage{british}{year-\textsc{num}} \foreignlanguage{british}{who-\textsc{sup}} \foreignlanguage{british}{go-\textsc{pst.3sg}}
            \glt \foreignlanguage{british}{`There also were 35 years old students.'}
            \glend 
    \item \label{ex-jed}
            \gll Ӕмӕ дын йед-мӕ куы ба-хӕццӕ дӕн… бын-мӕ...
            \foreignlanguage{british}{and} \foreignlanguage{british}{\textsc{2sg.dat}} \foreignlanguage{british}{\textsc{hes-all}} \foreignlanguage{british}{when} \foreignlanguage{british}{\textsc{pv}-mixed} \foreignlanguage{british}{be.\textsc{prs.1sg}} \foreignlanguage{british}{down-\textsc{all}}
            \glt \foreignlanguage{british}{`And when I came to... down there...'}
            \glend
\end{examples}

\selectlanguage{british}
The substantivation of pronominal adverbs is treated in a similar, albeit more restrictive, manner. Since these instances primarily involve the addition of case morphology — resulting in wordforms that continue to function adverbially or adjectivally — assigning a default Number=Sing feature is avoided as it may be misleading. Consequently, while such wordforms are assigned the PRON tag, we employ a heuristic whereby the Number feature is annotated only when represented by an overt marker (e.g., \foreignlanguage{russian}{\textit{афтӕтӕ}} in example \ref{ex-pradv}).
\selectlanguage{russian}
\begin{examples}
    \item \label{ex-pradv}
            \gll Ам дӕр афтӕ-т-ӕ нӕ уыд-ысты...
            \foreignlanguage{british}{DemProx.\textsc{ine.sg}} \foreignlanguage{british}{\textsc{add}} \foreignlanguage{british}{like.this-\textsc{pl-nom}} \foreignlanguage{british}{\textsc{neg}} \foreignlanguage{british}{be-\textsc{pst.3pl}}
            \glt \foreignlanguage{british}{`And it was not like this here...'}
            \glend 
\end{examples}

\selectlanguage{british}
\subsection{Additional cases}
The number of cases in Ossetic nominal paradigm ranges from nine to twelve \cite{belyaev:10}. The more recently grammaticalized cases — namely the directive and the recessive — are absent from classical grammars; however, following the mentioned study, we incorporate them into the nominal system. To accommodate these categories, we have introduced \textrm{Dir} and \textrm{Rcs} as values for the \textrm{Case} feature.

The status of the numerative as a distinct case, restricted to counted nouns in constructions with cardinal numerals, remains a subject of debate. This is primarily due to the homonymy of its marker (\foreignlanguage{russian}{\textit{-ы}}) with that of the genitive. Nevertheless, we identify at least two compelling arguments for its inclusion in the case paradigm. \par
First, in Digor -- a closely related dialect of Ossetic -- the marker of the corresponding form can co-occur with any case. This suggests that the numerative represents a separate grammatical function rather than a subfunction of the genitive. However, while Digor provides sufficient evidence to classify the numerative as an element of the number paradigm, Iron lacks such evidence, even though the need to distinguish it from the genitive persists. \par
Second, for pluralia tantum nouns, the numerative is homonymous with the nominative rather than the genitive. Consequently, we introduce \textrm{Num} as an additional value of the \textrm{Case} feature.

Regarding the numerative, specific disambiguation strategies are required. When a numeral construction functions as a direct object, we apply the principles of Differential Object Marking (DOM) for case assignment \cite{serdob:25}. Under these rules, human direct objects are typically genitive-marked, whereas inanimate objects appear in the nominative. Accordingly, in such contexts, we assign the genitive to human and the numerative to inanimate counted objects (e.g., animate \foreignlanguage{russian}{\textit{лӕппуйы}} in example \ref{ex-num-an} and inanimate \foreignlanguage{russian}{\textit{рӕнхъы}} in example \ref{ex-num-in}). Given that DOM patterns for non-human animates remain insufficiently defined, we adopt a heuristic approach, treating them similarly to inanimate nouns (e.g., \foreignlanguage{russian}{\textit{карчы}} in example \ref{ex-num-third}).

\selectlanguage{russian}
\begin{examples}
    \item \label{ex-num-an}
            \gll Уыцы дыууӕ лӕппу-йы с-хъомыл код-т-а
            \foreignlanguage{british}{DemDist} \foreignlanguage{british}{two} \foreignlanguage{british}{boy-\textsc{gen}} \foreignlanguage{british}{\textsc{pv}-adult} \foreignlanguage{british}{do-\textsc{tr-pst.3sg}}
            \glt \foreignlanguage{british}{`She brought up those two children'}
            \glend 
    \item \label{ex-num-in}
            \gll ...аст рӕнхъ-ы ны-ффыст-он.
            \foreignlanguage{british}{eight} \foreignlanguage{british}{line-\textsc{num}} \foreignlanguage{british}{\textsc{pv}-write-\textsc{pst.1sg}}
            \glt \foreignlanguage{british}{`I wrote about eight lines.'}
            \glend 
    \item \label{ex-num-third}
            \gll ...ӕмӕ иу-дӕс карч-ы а-ргӕвст-он, иуӕндӕс карч-ы
            \foreignlanguage{british}{and} \foreignlanguage{british}{one-ten} \foreignlanguage{british}{chicken-\textsc{num}} \foreignlanguage{british}{\textsc{pv}-slaughter-\textsc{pst.1sg}} \foreignlanguage{british}{eleven} \foreignlanguage{british}{chicken-\textsc{num}}
            \glt \foreignlanguage{british}{`...and slaughtered about a dozen chickens, eleven chickens.'}
            \glend 
\end{examples}

\selectlanguage{british}
\subsection{Non-finite verbal forms}
The extensive repertoire of non-finite verbal forms in Ossetic exceeds the inventory of values provided by the UD v2 specifications for the VerbForm feature. Consequently, we have introduced the additional values Inf2, Part2, Dest, and Red. The comprehensive mapping of all Ossetic non-finite forms to their respective VerbForm values is detailed in Table \ref{tab-nonf}, using the verb \foreignlanguage{russian}{\textit{дзурын}} (‘to speak’) as an illustrative example.

\begin{table}[ht!]
\centering
    \begin{tabular}{|l|l|l|l|}
        \hline
        Form & Value & Other features & Function \\
        \hline
        \foreignlanguage{russian}{\textit{дзур-ын}} & Inf & -- & Infinitive\\
        \foreignlanguage{russian}{\textit{дзур-ӕн}} & Inf2 & -- & Potential form (Infinitive II) \\
        \foreignlanguage{russian}{\textit{дзырд}} & Part & Tense=Past|Voice=Pass & Past participle (passive) \\
        \foreignlanguage{russian}{\textit{дзур-инаг}} & Part & Tense=Fut|Voice=Act/Pass & Future participle \\
        \foreignlanguage{russian}{\textit{дзур-ӕг}} & Part & Tense=Pres|Voice=Act & Present participle (active) \\
        \foreignlanguage{russian}{\textit{дзур-аг}} & Part2 & Tense=Pres|Voice=Act & Present participle II (active, habitual)\\
        \foreignlanguage{russian}{\textit{дзур-гӕ}} & Conv & -- & Converb \\
        \foreignlanguage{russian}{\textit{дзур-ӕггаг}} & Dest & -- & Destinative participle \\
        \foreignlanguage{russian}{\textit{ра-дзур-ба-дзур}} & Red & Preverb=Yes & Nominalization in complex predicates\\
        \hline
    \end{tabular}
    \caption{Non-finite verbal forms.}
    \label{tab-nonf}
\end{table}

\subsection{Prefixes}
In Ossetic, applying the Aspect feature to verbal forms would result in heterogeneous categories that do not align with the established tradition of Ossetic linguistics. Instead, we propose the feature Preverb=Yes for wordforms containing verbal prefixes, as these elements are the primary exponents of aspectual distinctions (e.g., \foreignlanguage{russian}{\textit{ӕрбахонон}} in example \ref{ex-pv}).

Furthermore, the elements \foreignlanguage{russian}{\textit{ӕд}} and \foreignlanguage{russian}{\textit{ӕнӕ}} function within texts as either nominal prefixes or independent prepositions (see \foreignlanguage{russian}{\textit{ӕнӕ}} in examples \ref{ex-pn} and \ref{ex-pp}). In instances of prefixation, we denote their presence by introducing the feature Prenoun=Yes for the respective wordforms.
\selectlanguage{russian}
\begin{examples}
    \item \label{ex-pv}
            \gll Ӕрба-хон-он дын ӕй?
            \foreignlanguage{british}{\textsc{pv}-call-\textsc{sbjv.2sg}} \foreignlanguage{british}{\textsc{2sg.dat}} \foreignlanguage{british}{\textsc{3sg.gen}}
            \glt \foreignlanguage{british}{`Shall I call her?'}
            \glend
    \item \label{ex-pn}
          \gll Адӕм ӕнӕ-хӕлд гъе уӕд уыд-ысты.
          \foreignlanguage{british}{person.\textsc{pl}} \foreignlanguage{british}{without-spoil.\textsc{ptcp.pst}} \foreignlanguage{british}{well} \foreignlanguage{british}{then} \foreignlanguage{british}{be-\textsc{pst.3pl}}
          \glt \foreignlanguage{british}{`That time people were unspoilt.'}
          \glend
    \item \label{ex-pp}
          \gll Ӕз та ӕнӕ очки-т-ӕй нӕ уыд-т-он бинаг фыст...
          \foreignlanguage{british}{\textsc{1sg}} \foreignlanguage{british}{\textsc{part}} \foreignlanguage{british}{without} \foreignlanguage{british}{glasses-\textsc{pl-abl}} \foreignlanguage{british}{\textsc{neg}} \foreignlanguage{british}{see-\textsc{tr-pst.1sg}} \foreignlanguage{british}{lower} \foreignlanguage{british}{write.\textsc{ptcp.pst}}
          \glt  \foreignlanguage{british}{`I didn't see the lower inscription without glasses...'}
          \glend
\end{examples}

\selectlanguage{british}
\section{Morphological analyzer creation}

In addition to the creation of a morphologically annotated corpus, the goal of our research is to release a morphological analyzer trained on the corpus data. Given the success of transfer learning approach and pretrained BERT-based models in morphological analysis (see \cite{kondratyuk-straka-2019-75}, \cite{straka-etal-2019-udpipe}, among others), we select BERT \cite{devlinetal2019} as the backbone model of our analyzer. Since there are no language models available for Ossetic, we train the BERT model on the standard task of masked language modeling and further finetune it on morphological labeling. The pretrained BERT model might be utilized for other Ossetic NLP tasks as well.

\label{sec-model}
\subsection{Unlabeled dataset}
We leverage unlabeled data from ONC to pre-train two BERT-based models. The corpus undergoes an automated preprocessing pipeline comprising the following stages:
\begin{enumerate}
\item NFC normalization.
\item Local deduplication via the removal of repeating strings within individual documents.
\item Extraction of HTML and JavaScript fragments using regular expressions.
\item Global deduplication to eliminate strings occurring more than 50 times across the entire corpus.
\item Document filtering based on a minimum length threshold of 200 characters.
\item Locality Sensitive Hashing (LSH) for near-duplicate detection and removal.
\end{enumerate}

The processed data is subsequently segmented into sentences using regular expressions and undergoes case normalization. Furthermore, sentences are excluded if they meet any of the following criteria:
\begin{itemize}
\item They consist exclusively of editorial or metadata.
\item They lack Cyrillic characters.
\item They contain no lowercase letters (all-caps).
\item They exhibit mixed Cyrillic and Latin scripts within a single token (excluding the Latin character \textit{æ}).
\end{itemize}
The resulting dataset comprises 1,060,693 sentences, partitioned into training and test sets with an 80:20 ratio (4:1).

\subsection{Pre-training}

We selected the following architectures as candidate models:
\begin{enumerate}
\item mBERT: Google’s multilingual BERT base (cased) \footnote{https://huggingface.co/google-bert/bert-base-multilingual-cased}.
\item Ossetic BERT: A version of mBERT featuring a custom embedding layer adapted to the target language, following the approach of \cite{kuratov2019adaptation}.
\end{enumerate}

To align the multilingual model with the character inventory of our training set, we augmented the standard mBERT tokenizer normalization with the following steps:
\begin{itemize}
\item The Cyrillic character \foreignlanguage{russian}{\textit{ӕ}} is mapped to its Latin counterpart, as the latter is the only form present in the mBERT vocabulary.
\item Various quotation mark variants (U+2018, U+2019, U+02BC, U+0027, U+00B4, U+0060) are normalized to the single character U+02BC. This uniformity is critical, as the quotation mark functions as a sandhi marker in Ossetic.
\end{itemize}

For the Ossetic BERT initialization, we first trained a WordPiece tokenizer on ONC data with a vocabulary size of 25,000 tokens. To preserve the knowledge from the original model, we initialized the new embedding matrix by tokenizing each new entry with the original mBERT tokenizer and averaging the resulting subtoken embeddings. The original mBERT embedding matrix was then replaced by the concatenated embeddings of the new Ossetic vocabulary.

Both models underwent fine-tuning on the Masked Language Modeling (MLM) objective for 5 epochs, with a batch size of 64 and a learning rate of $5\times10^{-5}$. MLM loss reached by each model is provided in Table \ref{tab-mlm}. These numbers are not directly comparable since model vocabularies are incompatible, which is illustrated by average token-to-word ratio in COT test set.

\begin{table}[ht!]
\begin{center}
\begin{tabular}{|l|cc|}
\hline & \bf MLM Loss & \bf Token-to-word Ratio \\ \hline
Multilingual & 0.6086 & 3.29 \\
Ossetic & 3.4311 & 1.21 \\
\hline
\end{tabular}
\end{center}
\caption{MLM results for multilingual and Ossetic models.}
\label{tab-mlm} 
\end{table}

\subsection{Fine-tuning}
Table \ref{tab-acc} summarizes the fine-tuning results over 10 epochs for multilingual model and 13 epochs for Ossetic model, with a batch size of 8 and a learning rate of $5\times10^{-5}$. These figures represent the average performance across five independent training runs. For each architecture, optimal hyperparameters were determined using the development set, and only the results corresponding to the best-performing configurations are reported.

\begin{table}[ht!]
\begin{center}
\begin{tabular}{|l|cc|}
\hline & \bf Accuracy & \bf Sentence Accuracy \\ \hline
Multilingual & 95.60 & 61.54 \\
Ossetic & 95.47 & 60.11 \\
\hline
\end{tabular}
\end{center}
\caption{Accuracy for multilingual and Ossetic models.}
\label{tab-acc} 
\end{table}

Since the performance gap between the two models was found to be statistically insignificant, the classifier based on multilingual BERT (mBERT) was selected as the final baseline. Detailed Precision, Recall, and F1-score metrics for this model are provided in Table \ref{tab-prf}. We report micro-, macro-, and weighted averages across all classes, as well as macro- and weighted averages aggregated by part-of-speech category. Each metric was calculated for each of the five training runs and subsequently averaged to ensure the robustness of the results.

\begin{table}[ht!]
\begin{center}
\begin{tabular}{|l|cccccc|}
\hline & \multicolumn{2}{c}{\bf Precision} & \multicolumn{2}{c}{\bf Recall} & \multicolumn{2}{c|}{\bf F1}  \\ 
& Macro & Weighted & Macro & Weighted & Macro & Weighted \\
    \hline
    ADJ & 0.78 & 0.91 & 0.81 & 0.92 & 0.79 & 0.91 \\
    ADP & 0.99 & 0.99 & 0.96 & 0.96 & 0.97 & 0.97 \\
    ADV & 0.84 & 0.95 & 0.88 & 0.97 & 0.86 & 0.96 \\
    AUX & 0.64 & 0.86 & 0.66 & 0.88 & 0.64 & 0.86 \\
    CCONJ & 0.99 & 0.99 & 0.99 & 0.99 & 0.99 & 0.99 \\
    DET & 0.89 & 0.96 & 0.87 & 0.93 & 0.88 & 0.94 \\
    INTJ & 0.98 & 0.97 & 0.98 & 0.97 & 0.98 & 0.97 \\
    NOUN & 0.78 & 0.93 & 0.78 & 0.94 & 0.78 & 0.93 \\
    NUM & 0.85 & 0.99 & 0.85 & 0.98 & 0.85 & 0.98 \\
    PART & 0.97 & 0.98 & 0.99 & 0.99 & 0.98 & 0.99 \\
    PRON & 0.81 & 0.92 & 0.82 & 0.92 & 0.80 & 0.92 \\
    PROPN & 0.44 & 0.95 & 0.50 & 0.96 & 0.46 & 0.96 \\
    PUNCT & 1.00 & 1.00 & 1.00 & 1.00 & 1.00 & 1.00 \\
    SCONJ & 0.93 & 0.86 & 0.91 & 0.82 & 0.92 & 0.84 \\
    VERB & 0.77 & 0.92 & 0.79 & 0.92 & 0.77 & 0.92 \\
    X & 0.87 & 0.87 & 0.66 & 0.66 & 0.75 & 0.75 \\
    \hline
    Micro Avg & \multicolumn{2}{c}{0.96} & \multicolumn{2}{c}{0.96} & \multicolumn{2}{c|}{0.96} \\
    Macro Avg & \multicolumn{2}{c}{0.77} & \multicolumn{2}{c}{0.78} & \multicolumn{2}{c|}{0.77} \\
    Weighted Avg & \multicolumn{2}{c}{0.95} & \multicolumn{2}{c}{0.96} & \multicolumn{2}{c|}{0.95} \\ 
\hline
\end{tabular}
\end{center}
\caption{Classification metrics for multilingual model.}
\label{tab-prf} 
\end{table}

\subsection{Homonymy resolution}
To provide a qualitative rather than purely quantitative evaluation of the model’s performance, we created a set of sentence pairs for principal instances of morphological homonymy in Ossetic. Each pair consists of similar sentences where a chosen target word requires different gold-standard tags depending on the context. The classifier’s task is to correctly resolve the homonymy in both instances. This test set is detailed in Appendix \ref{app-homon} with target words highlighted in blue. 
Table \ref{tab-homon} reports the model's predictions within these diagnostic contexts.

\begin{table}[ht!]
\begin{center}
\begin{tabular}{|l|ll|}
\hline 
& \bf Context I & \bf Context II \\ 
\hline
Adj/Noun & Adj & Noun \\
Adj/Adv & Adj & Adv \\
Noun/Adp & Noun & Noun \\
\textsc{1sg}/\textsc{inf} & \textsc{1sg} & \textsc{inf} \\
\textsc{gen}/\textsc{ine} & \textsc{gen} & \textsc{ine} \\
\textsc{gen}/\textsc{num} (inanimate) & \textsc{gen} & \textsc{num} \\
\textsc{gen}/\textsc{num} (humans) & \textsc{num} & \textsc{num} \\
\textsc{ine}/\textsc{prs.3sg} & \textsc{gen} & \textsc{gen} \\
DemDist.\textsc{nom}/DemDist.\textsc{gen} & DemDist.\textsc{nom} & DemDist.\textsc{gen} \\
\textsc{1pl.nom}/\textsc{1pl.gen} & \textsc{1pl.nom} & \textsc{1pl.gen} \\
\textsc{1sg.poss}/\textsc{1sg.gen} (enclitic) & \textsc{1sg.poss} & \textsc{1sg.gen} (enclitic) \\
\textsc{1sg.poss}/`and' & \textsc{1sg.poss} & `and' \\
\textsc{2sg.poss}/\textsc{2sg.gen} (enclitic) & \textsc{2sg.poss} & \textsc{2sg.gen} (enclitic) \\
\textsc{2sg.gen} (enclitic)/\textsc{be.prs.2sg} & \textsc{2sg.gen} (enclitic) & \textsc{be.prs.2sg} \\
\textsc{1pl.poss}/\textsc{1pl.gen} (enclitic) & \textsc{1pl.poss} & \textsc{neg} \\
\textsc{1pl.poss}/\textsc{neg} & \textsc{1pl.poss} & \textsc{neg} \\
\textsc{emph}/\textsc{neg} & \textsc{emph} & \textsc{neg} \\
\textsc{add}/\textsc{neg} & \textsc{add} & \textsc{neg} \\
\textsc{iter}/`one' & \textsc{iter} & `one' \\
\hline
\end{tabular}
\end{center}
\caption{Homonymy resolution for multilingual model.}
\label{tab-homon} 
\end{table}
Analysis of the results indicates that the most challenging instances of homonymy involve the following categories: Noun/Adp, \textsc{gen/num} for human-denoting nouns, \textsc{1pl.gen/neg} and \textsc{ine/prs.3sg}. Notably, the resolution of several instances of homonymy exhibits variability across different training runs. Furthermore, minor lexical variations within test examples can lead to inconsistent predictions for the target wordform. These phenomena may be attributed to the quantitative and lexical scarcity of COT, respectively. Such limitations could potentially be mitigated by augmenting the training set with more lexically diverse linguistic data.

\subsection{Errors examination}
A detailed examination of model errors on the test set allows for the identification of the following major error classes:
\begin{enumerate}
\item Inherent ambiguities: Cases where the context remains insufficient for definitive disambiguation.
\item Data imbalance: Failure to resolve homonymy for forms where one of the potential analyses is under-represented in the training set.
\item Orthographic confusion: Misidentification of orthographically similar wordforms. For instance, the form \foreignlanguage{russian}{\textit{мӕхицӕй}} (\textsc{1sg.refl.abl}) is mistaken for \foreignlanguage{russian}{\textit{мӕхицӕн}} (\textsc{1sg.refl.dat}) and is assigned a tag appropriate for the latter.
\item Long-range dependencies: Errors occurring in contexts where the cues necessary for correct analysis are positioned distantly within the sentence.
\end{enumerate}
Regarding specific linguistic categories, errors frequently occur in the identification of non-finite verbal forms, even when the part-of-speech tag and inflectional features are predicted correctly. Furthermore, the model occasionally fails to distinguish between the auxiliary and lexical uses of verbs.

\section{Limitations}
\label{sec-lim}
Notably, both the development and validation sets contain several feature combinations that are absent from the training set. Furthermore, a subset of theoretically possible combinations does not appear in any of the sets. While the challenge of identifying these previously unseen classes is typically addressed through a multi-task learning approach (e.g., \cite{inoue:17,straka-etal-2019-udpipe}) by predicting the values of each morphological feature (e.g., Case) separately. However, we observed that the performance of a multi-task model on our dataset did not significantly exceed that of the baseline classifier described in Section \ref{sec-model}. Consequently, we prioritize the latter for its simplicity and computational efficiency.

As noted in Section \ref{sec-model}, COT exhibits limited lexical diversity. Moreover, the syntactic structures in the majority of sentences are either notably simple or characteristic of oral discourse. These data-specific properties may result in a performance degradation when the model, trained on our corpus, is applied to out-of-domain data or more complex literary registers. We leave the extension of our analyzer to wriiten data for future work.

\section{Concluding remarks}
\label{sec-concl}
In this paper, we have presented the first Ossetic corpus with morphological annotation adhering to the UD v2 schema. This resource is suitable for training neural morphological taggers, as well as for linguistic research and educational applications, as its annotation captures the majority of morphological features and distinctions inherent to the language. Furthermore, we provided a baseline model that serves both as a benchmark for evaluating dataset complexity and as an initial neural morphological tagger for Ossetic. The dataset and the model are freely available \footnote{https://github.com/ania3000/Ossetic-COT} \footnote{https://huggingface.co/ossetic-encoders/ossbert-morph}.

The primary direction for future research involves addressing the current quantitative and qualitative data scarcity. We aim to mitigate these limitations by augmenting the corpus with more diverse literary data from ONC, thereby enhancing the model's robustness across different registers.

\section*{Acknowledgements}
The authors are grateful to Oleg Belyaev for inspiring this research, for careful assistance in developing the annotation guidelines and for providing unlabeled training data.

\bibliography{dialogue.bib}

@inproceedings{inoue:17,
    title = "{J}oint {P}rediction of {M}orphosyntactic {C}ategories for {F}ine-{G}rained {A}rabic {P}art-of-{S}peech {T}agging {E}xploiting {T}ag {D}ictionary {I}nformation",
    author = "Inoue, Go  and
      Shindo, Hiroyuki  and
      Matsumoto, Yuji",
    editor = "Levy, Roger  and
      Specia, Lucia",
    booktitle = "Proceedings of the 21st Conference on Computational Natural Language Learning ({C}o{NLL} 2017)",
    year = "2017",
    address = "Vancouver, Canada",
    publisher = "Association for Computational Linguistics",
    pages = "421--431",
}

@inproceedings{kondratyuk-straka-2019-75,
    title = "75 {L}anguages, 1 {M}odel: Parsing {U}niversal {D}ependencies {U}niversally",
    author = "Kondratyuk, Dan  and
      Straka, Milan",
    editor = "Inui, Kentaro  and
      Jiang, Jing  and
      Ng, Vincent  and
      Wan, Xiaojun",
    booktitle = "Proceedings of the 2019 Conference on Empirical Methods in Natural Language Processing and the 9th International Joint Conference on Natural Language Processing (EMNLP-IJCNLP)",
    year = "2019",
    address = "Hong Kong, China",
    publisher = "Association for Computational Linguistics",
    url = "https://aclanthology.org/D19-1279/",
    doi = "10.18653/v1/D19-1279",
    pages = "2779--2795"
}

@inproceedings{straka-etal-2019-udpipe,
    title = "{UDP}ipe at {SIGMORPHON} 2019: {C}ontextualized {E}mbeddings, {R}egularization with {M}orphological {C}ategories, {C}orpora {M}erging",
    author = "Straka, Milan  and
      Strakov{\'a}, Jana  and
      Hajic, Jan",
    editor = "Nicolai, Garrett  and
      Cotterell, Ryan",
    booktitle = "Proceedings of the 16th Workshop on Computational Research in Phonetics, Phonology, and Morphology",
    year = "2019",
    address = "Florence, Italy",
    publisher = "Association for Computational Linguistics",
    url = "https://aclanthology.org/W19-4212/",
    doi = "10.18653/v1/W19-4212",
    pages = "95--103"
}

@misc{kuratov2019adaptation,
  title={Adaptation of {D}eep {B}idirectional {M}ultilingual {T}ransformers for {R}ussian {L}anguage},
  author={Kuratov, Yuri and Arkhipov, Mikhail},
  year={2019}
}

@book{belyaev:14,
    author  = {Belyaev, Oleg},
    title   = {Korreljativnaja konstrukcija v osetinskom jazyke v tipologicheskom osveschenii},
    year    = "2014",
    publisher = {Lomonosov Moscow State University},
    address = {Moscow}
}

@inproceedings{archangelsky:12,
    title = "The {C}reation of {L}arge-{S}cale {A}nnotated {C}orpora of {M}inority {L}anguages using {U}ni{P}arser and the {EANC} platform",
    author = "Arkhangelskiy, Timofey  and
      Belyaev, Oleg  and
      Vydrin, Arseniy",
    editor = "Kay, Martin  and
      Boitet, Christian",
    booktitle = "Proceedings of {COLING} 2012: Posters",
    year = "2012",
    address = "Mumbai, India",
    publisher = "The COLING 2012 Organizing Committee",
    pages = "83--92"
}

@book{abaev:64,
    author  = {Abaev, Vasilij},
    title   = {A grammatical sketch of {O}ssetic},
    year    = "1964",
    publisher = {Mouton},
    address = {The Hague}
}

@article{belyaev:10,
	author = {Belyaev, Oleg},
	year = "2010",
    title = {{E}volution of {C}ase in {O}ssetic},
	journal = {Iran and the Caucasus},
	volume = "14",
	number = "2",
	pages = "287--322",
}

@article{serdob:25,
	author = {Serdobolskaya, Natalia and Tuzhik, Olga},
	year = "2025",
    title = {{D}ifferential {O}bject {M}arking in {M}odern {O}ssetic: {R}eferential {P}roperties and {A}nimacy},
	journal = {Tomsk Journal of Linguistics and Anthropology},
	volume = "48",
	number = "2",
	pages = "91--107",
}

@inproceedings{devlinetal2019,
    title = "{BERT}: {P}re-training of {D}eep {B}idirectional {T}ransformers for {L}anguage {U}nderstanding",
    author = "Devlin, Jacob  and
      Chang, Ming-Wei  and
      Lee, Kenton  and
      Toutanova, Kristina",
    editor = "Burstein, Jill  and
      Doran, Christy  and
      Solorio, Thamar",
    booktitle = "Proceedings of the 2019 Conference of the North {A}merican Chapter of the Association for Computational Linguistics: Human Language Technologies, Volume 1 (Long and Short Papers)",
    year = "2019",
    address = "Minneapolis, Minnesota",
    publisher = "Association for Computational Linguistics",
    pages = "4171--4186"
}

@inproceedings{nivre:2020,
    title = "{U}niversal {D}ependencies v2: An {E}vergrowing {M}ultilingual {T}reebank {C}ollection",
    author = "Nivre, Joakim  and
      de Marneffe, Marie-Catherine  and
      Ginter, Filip  and
      Haji{\v{c}}, Jan  and
      Manning, Christopher D.  and
      Pyysalo, Sampo  and
      Schuster, Sebastian  and
      Tyers, Francis  and
      Zeman, Daniel",
    editor = "Calzolari, Nicoletta  and
      B{\'e}chet, Fr{\'e}d{\'e}ric  and
      Blache, Philippe  and
      Choukri, Khalid  and
      Cieri, Christopher  and
      Declerck, Thierry  and
      Goggi, Sara  and
      Isahara, Hitoshi  and
      Maegaard, Bente  and
      Mariani, Joseph  and
      Mazo, H{\'e}l{\`e}ne  and
      Moreno, Asuncion  and
      Odijk, Jan  and
      Piperidis, Stelios",
    booktitle = "Proceedings of the Twelfth Language Resources and Evaluation Conference",
    year = "2020",
    address = "Marseille, France",
    publisher = "European Language Resources Association",
    url = "https://aclanthology.org/2020.lrec-1.497/",
    pages = "4034--4043",
}
\bibliographystyle{dialogue}

\appendix
\section{Appendix \ref{app-gen}. Frequency distribution of subgenres in COT}
\label{app-gen}
\begin{table}[ht!]
\begin{tabular}{|l|ccc|}
\hline 
& \textbf{Train} & \textbf{Development} & \textbf{Test}\\
\hline
Radio & 253 (6.15\%) & 37 (6.79\%) & 29 (5.32\%)\\
Expedition & 4111 (93.85\%) & 508 (93.21\%) & 516 (94.68\%) \\
\hline
\end{tabular}
\end{table}

\section{Appendix \ref{app-freq}. Frequency distribution for 20 most frequent tags}
\label{app-freq}
\begin{table}[ht!]
\begin{longtable}{|l|ccc|}
\hline 
& \textbf{Train} & \textbf{Development} & \textbf{Test}\\
\hline
PUNCT & 14913 (25.58\%) & 1865 (24.92\%) & 1824 (25.16\%) \\
NOUN,Case=Nom|Number=Sing & 3269 (5.61\%) & 450 (6.01\%) & 444 (6.12\%) \\
PART & 3236 (5.55\%) & 419 (5.60\%) & 394 (5.43\%) \\
CCONJ & 2794 (4.79\%) & 351 (4.69\%) & 355 (4.90\%) \\
PROPN,Case=Nom & 1888 (3.24\%) & 253 (3.38\%) & 277 (3.82\%) \\
ADV & 1027 (1.76\%) & 109 (1.46\%) & 127 (1.75\%) \\
ADV,PronType=Int & 1009 (1.73\%) & 112 (1.50\%) & 116 (1.60\%) \\
NOUN,Case=Nom|Number=Plur & 1003 (1.72\%) & 126 (1.68\%) & 140 (1.93\%) \\
ADV,PronType=Dem & 999 (1.71\%) & 123 (1.64\%) & 139 (1.92\%) \\
PART,Polarity=Neg & 988 (1.69\%) & 111 (1.48\%) & 124 (1.71\%) \\
ADJ & 882 (1.51\%) & 120 (1.60\%) & 104 (1.43\%) \\
ADV,Deixis=Remt|PronType=Dem & 851 (1.46\%) & 132 (1.76\%) & 111 (1.53\%) \\
VERB,Mood=Ind|Number=Sing| & & & \\
Person=3|Tense=Pres|VerbForm=Fin & 830 (1.42\%) & 100 (1.34\%) & 85 (1.17\%) \\
VERB,Mood=Ind|Number=Sing| & & & \\
Person=3|Preverb=Yes|Tense=Past|VerbForm=Fin & 749 (1.28\%) & 95 (1.27\%) & 87 (1.20\%) \\
INTJ & 714 (1.22\%) & 113 (1.51\%) & 132 (1.82\%) \\
ADV,Deixis=Prox|PronType=Dem & 701 (1.20\%) & 77 (1.03\%) & 92 (1.27\%) \\
VERB,Mood=Ind|Number=Sing| & & & \\
Person=3|Tense=Past|VerbForm=Fin & 673 (1.15\%) & 96 (1.28\%) & 98 (1.35\%) \\
PRON,Case=Gen|Number=Sing|Person=3|PronType=Prs & 643 (1.10\%) & 82 (1.10\%) & 88 (1.21\%) \\
NUM,Case=Nom|NumType=Card & 610 (1.05\%) & 63 (0.84\%) & 81 (1.12\%) \\
NOUN,Case=Gen|Number=Sing & 590 (1.01\%) & 95 (1.27\%) & 66 (0.91\%) \\
\hline
\end{longtable}
\end{table}

\section{Appendix \ref{app-homon}. Sentence set for homonymy resolution}
\label{app-homon}
\selectlanguage{russian}
\begin{table}[ht!]
\begin{tabular}{|l|ll|}
\hline 
& \foreignlanguage{english}{\bf Context I} & \foreignlanguage{english}{\bf Context II} \\ 
\hline
\foreignlanguage{english}{Adj/Noun} & {\color{blue}Зӕронд} ус ӕрбацыди. & {\color{blue}Зӕронд} ӕрбацыди. \\
\foreignlanguage{english}{Adj/Adv} & Уый {\color{blue}рӕсугъд} чызгимӕ кафы. & Уый {\color{blue}рӕсугъд} кафы. \\
\foreignlanguage{english}{Noun/Adp} & Лӕппу хӕдзары {\color{blue}сӕр} уыны. & Лӕппу хӕдзары {\color{blue}сӕр} бады. \\
\foreignlanguage{english}{\textsc{1sg}/\textsc{inf}} & Ӕз {\color{blue}кусын}. & Ӕз {\color{blue}кусын} байдыдтон. \\
\foreignlanguage{english}{\textsc{gen}/\textsc{ine}} & Залинӕ {\color{blue}хӕдзары} мидӕг уыди. & Залинӕ {\color{blue}хӕдзары} уыди. \\
\foreignlanguage{english}{\textsc{gen}/\textsc{num} (inanimate)} & Залинӕ {\color{blue}хӕдзары} хицау федта. & Залинӕ дыууӕ {\color{blue}хӕздары} федта. \\
\foreignlanguage{english}{\textsc{gen}/\textsc{num} (humans)} & Дыууӕ {\color{blue}лӕджы} федтон. & Дыууӕ {\color{blue}лӕджы} ӕрбацыдысты. \\
\foreignlanguage{english}{\textsc{ine}/\textsc{prs.3sg}} & Мах {\color{blue}хъӕуы} барӕгбон кӕнӕм. & Мах {\color{blue}хъӕуы} барӕгбон. \\
\foreignlanguage{english}{DemDist.\textsc{nom}/DemDist.\textsc{gen}} & {\color{blue}Уый} ӕрбацыдис. & {\color{blue}Уый} фӕстӕ ӕрбацыдтӕн. \\
\foreignlanguage{english}{\textsc{1pl.nom}/\textsc{1pl.gen}} & {\color{blue}Мах} уым цӕрӕм. & {\color{blue}Мах} бинонтӕ уым цӕрынц. \\
\foreignlanguage{english}{\textsc{1sg.poss}/\textsc{1sg.gen} (enclitic)} & Уый {\color{blue}мӕ} мады федта. & Уый {\color{blue}мӕ} федта.  \\
\foreignlanguage{english}{\textsc{1sg.poss}/`and'} & {\color{blue}Мӕ} чызг уым ахуыр кодта. & Лӕппу {\color{blue}мӕ} чызг уым ахуыр кодтой. \\
\foreignlanguage{english}{\textsc{2sg.poss}/\textsc{2sg.gen} (enclitic)} & Уый {\color{blue}дӕ} фыды федта. & Уый {\color{blue}дӕ} федта. \\
\foreignlanguage{english}{\textsc{2sg.gen} (enclitic)/be.\textsc{prs.2sg}} & Ирон лӕг {\color{blue}дӕ} федта. & Ирон лӕг {\color{blue}дӕ}. \\
\foreignlanguage{english}{\textsc{1pl.poss}/\textsc{1pl.gen} (enclitic)} & Уый {\color{blue}нӕ} чызджы федта. & Уый {\color{blue}нӕ} федта. \\
\foreignlanguage{english}{\textsc{1pl.poss}/\textsc{neg}} & Уый {\color{blue}нӕ} хӕдзар федта. & Уый хӕдзар {\color{blue}нӕ} федта. \\
\foreignlanguage{english}{\textsc{emph}/\textsc{neg}} & Радзур-{\color{blue}ма} исты. & {\color{blue}Ма} дзур. \\
\foreignlanguage{english}{\textsc{add}/\textsc{neg}} & Хъӕумӕ {\color{blue}ма} цыдтӕн. & Хъӕумӕ {\color{blue}ма} цӕу. \\
\foreignlanguage{english}{\textsc{iter}/`one'} & Ӕмӕ-{\color{blue}иу} лӕг ӕрбацыдис. & Ӕмӕ {\color{blue}иу} лӕг ӕрбацыдис. \\
\hline
\end{tabular}
\end{table}

\end{document}